\pdfoutput=1
\documentclass[11pt]{article}
\usepackage{emnlp2021}
\usepackage{times}
\usepackage{latexsym}
\usepackage{amsmath}
\usepackage{tikz}
\usepackage{subcaption}
\usepackage{hyperref}
\usepackage[T1]{fontenc}
\usepackage[utf8]{inputenc}
\usepackage{booktabs}
\usepackage{multicol, caption, multirow, hhline}
\usepackage{gb4e}
\noautomath
\usepackage{xspace}
\usepackage{appendix}
\usepackage{ amssymb }
\usepackage[normalem]{ulem}
\usepackage{tikz-dependency}
\usepackage{microtype}
\usepackage{hyperref}
\let\svthefootnote\thefootnote
\newcommand\freefootnote[1]{%
  \let\thefootnote\relax%
  \footnotetext{#1}%
  \let\thefootnote\svthefootnote%
}

\newcommand{\unmaskedmodel}{\textbf{M}$_\text{u}$\xspace}
\newcommand{\maskedmodel}{\textbf{M}$_\text{m}$\xspace}

\title{Does referent predictability affect the choice of referential form? \\ A computational approach using masked coreference resolution}

\author{Laura Aina$^{\bf{1*}}$, Xixian Liao$^{\bf{1*}}$, Gemma Boleda$^{\bf{1,2}}$ \and Matthijs Westera$^{\bf{3}}$ \\
$^1$Department of Translation and Language Sciences, Universitat Pompeu Fabra \\
$^2$Catalan Institution for Research and Advanced Studies - ICREA\\
$^3$Leiden University Centre for Linguistics, Universiteit Leiden \\
\texttt {\{laura.aina,xixian.liao,gemma.boleda\}@upf.edu} \\
\texttt {matthijs.westera@gmail.com}}

\begin{document}
\maketitle

\begin{abstract}
\freefootnote{* First two authors contributed equally.}
It is often posited that more predictable parts of a speaker's meaning tend to be made less explicit, for instance using shorter, less informative words.
Studying these dynamics in the domain of referring expressions has proven difficult, with existing studies, both psycholinguistic and corpus-based, providing contradictory results.
We test the hypothesis that speakers produce less informative referring expressions (e.g., pronouns vs.~full noun phrases) 
when the context is more informative about the referent, using novel computational estimates of referent predictability. 
We obtain these estimates training an existing coreference resolution system for English on a new task, masked coreference resolution, giving us a probability distribution over referents that is conditioned on the context but not the referring expression.
The resulting system retains standard coreference resolution performance while yielding a better estimate of human-derived referent predictability than previous attempts.
A statistical analysis of the relationship between model output and mention form supports the hypothesis that predictability affects the form of a mention, both its morphosyntactic type and its length.
\end{abstract}

\section{Introduction}
\label{sec:introduction}

A long-standing hypothesis in linguistics relates predictability to linguistic form: the more predictable parts of a speaker's meaning tend to be communicated with shorter, less informative words and be pronounced more quickly and flatly \citep{ferrer2003least,aylett2004smooth,levy2007speakers,piantadosi2011word}.
This is posited to result from a trade-off between clarity and cost:
when the context already clearly conveys certain information, making it explicit would be inefficient.
In the domain of referring expressions, or \emph{mentions}, the prediction is that pronouns such as ``they'' and ``her'' are used for more predictable referents while proper names such as ``Kamala Harris'' or full noun phrases such as ``my child's teacher'' are required for less predictable referents \citep{tily2009refer}.
The aim of this paper is to test this prediction using a novel, computational estimate of referent predictability.

Existing work on the relation between predictability and mention form paints a mixed picture (Section~\ref{sec:relatedwork}).
A set of studies in psycholinguistics used controlled stimuli designed to manipulate predictability \citep{arnold2001effect, rohde2014grammatical}.
An alternative approach uses naturally occurring corpus data to elicit judgments from human subjects \citep{tily2009refer, modi2017modeling}.
Neither approach, however, scales well.
We instead obtain predictability estimates from a coreference system (trained on English data); if successful, this strategy would allow to test the hypothesis in a wider set of contexts than in psycholinguistic experiments.

\begin{figure*}[hbt]
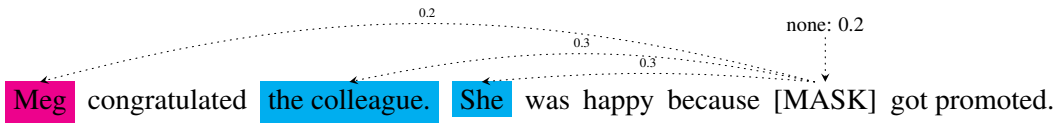


\begin{center}

 \begin{dependency}[theme = simple]
\begin{deptext}
\colorbox{magenta}{Meg} \&  congratulated  \& \colorbox{cyan}{the colleague.} \&  \colorbox{cyan}{She} \& was \& happy  \& because \& {[MASK]} \& got promoted. \\
   \end{deptext}
   \depedge[arc angle = 15, dotted, font=\normalsize]{8}{1}{0.2}
   \depedge[arc angle = 12, dotted,font=\normalsize]{8}{3}{0.3}
   \deproot[edge unit distance=1.2ex, dotted,font=\normalsize]{8}{none: 0.2}
   \depedge[arc angle = 5, dotted,font=\normalsize]{8}{4}{0.3}
 \end{dependency}
 \end{center}
 
 \small

\caption{An example of deriving referent probabilities from masked coreference resolution predictions.}
\label{fig:approach}
\end{figure*}

This paper follows the long tradition of using computational models trained on large amounts of data as proxies for different aspects of human cognition; in particular, it extends to the referential level previous research that uses computational models to obtain predictability scores \citep{hale2001probabilistic, smith2013effect}.
Standard coreference models, however, cannot be directly used to model predictability, because they are trained with access to both the context of a referring expression and the expression itself.
Instead, we aim at obtaining predictability scores that are only conditioned on the context, following the definition of predictability used in psycholinguistics.
To this end, we minimally modify a state-of-the-art coreference system \citep{joshi2020spanbert} to also carry out what we call \emph{masked coreference resolution} (Figure~\ref{fig:approach}): computing referent probabilities without observing the target mention.
We show that the resulting model retains standard coreference resolution performance, while yielding a better estimate of human-derived referent predictability than previous attempts.

Statistical analysis of the relationship between model scores and mention form suggests that predictability is indeed linked to referential form. 
Low referent surprisal tends to be associated with pronouns (as opposed to proper names and full noun phrases) and in general shorter expressions.
When controlling for shallow cues that modulate the salience of an entity (e.g., recency, frequency), surprisal no longer differentiates pronouns from proper names (while still favoring longer noun phrases). 
This may be read as supporting the hypothesis that it is primarily mention length that is at stake, rather than morphosyntactic category
, but it may also be due to stylistic factors in the data we use (OntoNotes, \citealt{weischedel2013ontonotes}).\footnote{
    We make the code used to carry out this study available at \url{https://github.com/amore-upf/masked-coreference}.
}

\section{Related work}
\label{sec:relatedwork}

\paragraph{Predictability and referential form.} Traditional approaches to discourse anaphora posit that the \textit{salience} of referents determines the choice of referring expression (a.o., \citealp{ariel1990accessing,gundel1993cognitive}).
Salience, however, is notoriously hard to characterize and operationalize.
Later work has therefore focused on \textit{predictability}, that is, how expected a specific referent is at a certain point in the discourse  (see \citealt{arnold2019people} for an overview).
This notion can be more easily operationalized to gather judgements from human subjects.
Previous work has done so using cloze tasks and discourse continuation experiments.

Some of this work found that more pronouns were produced for more predictable referents, suggesting an addressee-oriented strategy for pronoun production (e.g., \citealp{arnold2001effect}). 
Other work instead reported the same pronominalization rates in contexts that favored different referents, and argued for a mismatch between speaker and listener strategies (e.g.,\ \citealp{rohde2014grammatical, mayol2018asymmetries}). 
While the aforementioned psycholinguistic studies used tightly controlled stimuli, researchers also reported contradictory results when looking at cloze task responses in corpus data. \citet{tily2009refer} considered newspaper texts from the OntoNotes corpus and found that pronouns and proper names were favoured over full NPs when subjects were able to guess the upcoming referent.
By contrast, \citet{modi2017modeling} did not find this effect on narrative texts in the InScript corpus \citep{modi2017inscript}.

\paragraph{Computational estimates of predictability.}
Probabilistic language models have been commonly adopted to study expectation-based human language comprehension \citep{armeni2017probabilistic}.
Predictability scores obtained from language models have been shown to correlate with measures of cognitive cost at the lexical and the syntactic levels \citep{smith2013effect,frank2013word}. 
In this work, predictability is typically measured with solely the preceding context in computational psycholinguistics (e.g., \citealp{ levy2008expectation}).
However, more recent work has also looked at predictability calculated based on both the previous and following contexts using bidirectional networks, like we do: 
\citet{pimentel2020speakers} used surprisal calculated from a cloze language model which takes both left and right pieces of context, as the operationalization of word predictability. They studied the trade-off between clarity and cost and reported a tendency for ambiguous words to appear in highly informative contexts.

Previous work also used computational estimates of referent predictability. In \citet{orita2015discourse}, they were used to explain referential choice as the result of an addressee-oriented strategy. 
Their measures of predictability, however, were based on simple features like frequency or recency. 
\citet{modi2017modeling} built upcoming referent prediction models combining shallow linguistic features and script knowledge. 
This approach allowed them to disentangle the role of linguistic and common-sense knowledge, respectively, on human referent predictions.
More recent research assessed the ability of autoregressive language models to mimick referential expectation biases that humans have shown in the psycholinguistic literature (e.g., \citealp{upadhye2020predicting}).
\citet{davis2021uncovering} extended this assessment to non-autoregressive models, like the ones we use here, and reported results consistent with prior work in autoregressive models, showing that these models exhibited biases in line with existing evidence on human behavior, at least for English.

\paragraph{Automatic coreference resolution.}
The goal of a standard coreference resolution system is to group mentions in a text according to the real-world entity they refer to \citep{pradhan2012conll}.
Several deep learning approaches have been proposed in the NLP literature, such as cluster-ranking \citep{clark2016improving} or span-ranking architectures \citep{lee2017end}.
We focus on span-ranking models, which output, for each mention, a probability distribution over its potential antecedents.

We rely on an existing state-of-the-art `end-to-end' coreference resolution system based on the SpanBERT language model \citep{joshi2020spanbert}, henceforth SpanBERT-coref.
It builds directly on the coreference systems of \citet{joshi2019bert} and \citet{lee2018higher}, the main innovation being its reliance on SpanBERT in place of (respectively) BERT \citep{devlin2018bert} and ELMo \citep{peters2018elmo}.
We give more details about the system in Section~\ref{sec:method}.

SpanBERT and BERT are transformer encoders pretrained on masked language modeling \citep{devlin2018bert}:
a percentage of tokens is \textit{masked} -- substituted with a special token \texttt{[MASK]}; the model has to predict these tokens based on their surrounding context.
For training SpanBERT, random contiguous sets of tokens --\emph{spans}-- are masked, and an additional Span Boundary Objective encourages meaningful span representations at the span's boundary tokens.\footnote{In addition, the Next Sentence Prediction task used in BERT is dropped in SpanBERT, which allows the latter to be trained on larger contiguous segments of text.}

\section{Methods}
\label{sec:method}

\paragraph{The SpanBERT-coref system.}

We use the best system from \citet{joshi2020spanbert}, which is based on SpanBERT-base (12 hidden layers of 768 units, with 12 attention heads).
We make no changes to its architecture, only to the data on which it is trained, by masking some mentions.

In SpanBERT-coref, each span of text is represented by a fixed-length vector computed from SpanBERT token representations, obtained considering a surrounding context window of maximum 384 tokens.\footnote{A span representation is the concatenation of its start token representation, end token representation, and a weighted sum (attention) over all of its token representations.}
From span representations, a mention score is computed for each span ($s_m$: how likely it is to be a mention), and a compatibility score is computed for each pair of spans ($s_a$: how likely it is that they corefer).
These scores are aggregated into a score $s$ for each pair of spans (Eq.~\ref{scores}).
For each mention, a probability distribution over its candidate antecedents (previous mentions and `none') is then derived, determining coreference links (Eq. \ref{scores_to_probabilities}). 
The complete, end-to-end system is trained (and the underlying language model finetuned) on the English portion of the coreference-annotated OntoNotes 5.0 dataset \citep{weischedel2013ontonotes}, which we also use.
\begin{align}
    s(x, y) = s_m(x) + s_m(y) +  s_a(x, y) \label{scores} \\ 
    P(\text{antecedent}_x = y) = \frac{e^{s(x,y)}}{\sum_{i \in \text{candidate}_{x}} e^{s(x,i)}} \label{scores_to_probabilities}
\end{align}

\paragraph{Training on masked coreference.}

We model entity predictability in terms of a probability distribution over entities given a masked mention.
The probability that a mention $x$ refers to the entity $e$ -- $P(E_x = e)$ -- can be computed as the sum of the antecedent probabilities of the mentions of $e$ in the previous discourse ($M_e$):
\begin{align}
    P(E_x = e) = \sum_{i \in M_e} P (\text{antecedent}_x = i) \label{prob_entity}
\end{align}
\noindent However, in SpanBERT-coref this probability is conditioned both on the mention and its context (the model has observed both to compute a prediction), whereas we need a distribution that is only conditioned on the context.

To achieve this, we introduce \textit{masked coreference resolution}, the task of determining the antecedent of a mention that has been replaced by an uninformative token \texttt{[MASK]}.
The task, inspired on masked language modeling \citep{devlin2018bert}, is illustrated in Figure \ref{fig:approach}.
Note that SpanBERT-coref can be directly used for masked coreference, since its vocabulary already includes the \texttt{[MASK]} token.
However, since the system was not trained in this setup, its predictions are not optimal, as we show in Section~\ref{sec:maskedvsunmasked}.
Therefore, we train a new instance of SpanBERT-coref adapted to our purposes -- \maskedmodel.
To train \maskedmodel, we mask a random sample of mentions in each document by replacing them by a single \texttt{[MASK]} token.%
\footnote{
    Masked spans are re-sampled at every epoch.
    [MASK] replaces the entire span of each mention, independently of its length.
    We verified that the use of one [MASK] token did not bias \maskedmodel to expect single-token mentions such as pronouns; see Figure \ref{fig:masked_accuracy_three_masks} in Appendix.}
The percentage of mentions masked is a hyperparameter (we test 5\%-40\%).
Note that, this way, the model is optimized to identify the correct antecedents of both masked and unmasked mentions, such that it retains standard coreference capabilities (see Section~\ref{sec:maskedvsunmasked}).

\paragraph{Evaluation.}

To evaluate general coreference performance, following the CoNLL-2012 shared task  \citep{pradhan2012conll}, we report the averages of the MUC, B$^3$ and CEAF metrics in precision, recall and F1, respectively.
These metrics focus on the quality of the induced clusters of mentions (i.e., coreference chains) compared to the gold ones.

When using \maskedmodel to model predictability, however, we only care about antecedent assignments, not overall clustering structure.
Therefore, we also evaluate the models on the task of \textit{antecedent prediction}.
We compute antecedent precision, recall and F1, where a model's prediction for a mention is correct if it assigns the largest probability to a true antecedent (an antecedent belonging to the correct mention cluster), or to none if it is the first mention of an entity.
We use F1 on antecedent prediction as the criterion for model selection during hyperparameter tuning.

To obtain predictions at test time, given a masked mention, we derive a probability distribution over its \emph{unmasked} antecedents (Figure \ref{fig:approach}).
Thus, when computing masked predictions we must avoid interference between masked mentions.
At the same time, for computational efficiency, we want to avoid masking only one mention per document at a time.
We solve this by computing, for each document, a partition of the mentions, where each subset is \textit{maskable} if, for each mention, none of its antecedents nor surrounding tokens (50 on either side) are masked.
We generate one version of each document for each subset of masked mentions. 
We compute predictions for each document version separately, and collect antecedent assignments for the masked mentions, thereby obtaining masked predictions for each mention in the document.%
\footnote{For hyperparameter tuning on development data, we use a faster but more coarse-grained method, described in Appendix \ref{appendix_a}.}

\paragraph{Using gold mention boundaries.} As mentioned earlier, SpanBERT-coref is jointly trained to detect mentions and to identify coreference links between them.
We are interested only in the latter task, for which the challenge of mention detection only adds noise.
Therefore, for analysis (not during training) we use gold mentions as the only candidate spans to consider; mention scores are set to 0, nullifying their contribution to model predictions (i.e., $s(i, j) = s_a(i, j)$; Eq. \ref{scores}).

\paragraph{Context.} 
Our setup differs from that of \citet{tily2009refer} and \citet{modi2017modeling} in that their human participants were given only the context preceding the mention, while our model is given context on both sides, in line with the bidirectional nature of BERT, SpanBERT and the state of the art in coreference resolution.
Thus, the notion of referent predictability we model is to be understood not in the sense of anticipation in time, but in informational terms (in line with \citealt{levy2007speakers}): how much information the context provides vs.\ how much information the mention needs to provide for its intended referent to be understood.
This is not necessarily less cognitively plausible than a left-context-only setting: Humans take right-hand context into account when interpreting referring expressions \citep{deemter1990forward, song2020forward}, Indeed, this could provide crucial disambiguating information, as shown in the following example:

\begin{exe}
\ex \colorbox{magenta}{Ann} scolded \colorbox{cyan}{her daughter}, because \label{ex:ref_right_context}
\begin{itemize}
    \item[a)] \colorbox{cyan}{\textbf{she}} was not behaving. \vspace{-0.1cm}
    \item[b)] \colorbox{magenta}{\textbf{she}} was not happy with her behavior. 
\end{itemize}
\end{exe}

We leave the exploration of different kinds of context to future work.

\section{Evaluation}
\label{sec:maskedvsunmasked}

\begin{table*}[h]
    \centering
    \begin{tabular}{llcccccccp{0.3cm}ccc} \toprule
        & & \multicolumn{6}{c}{\textsc{Unmasked mentions}} & &  \multicolumn{3}{c}{\textsc{Masked mentions}} \\   \toprule
        & & \multicolumn{3}{c}{\textsc{Coreference}} & \multicolumn{3}{c}{\textsc{Antecedent}} & & \multicolumn{3}{c}{\textsc{Antecedent}} \\
        & boundaries & P & R & F1 & P & R & \ \ \ \ \ F1 \ \ \ \ \ & & P & R & F1 \\ \toprule
       \unmaskedmodel & predicted  & .78	 & .77	 & .77 & .86  & .82 & .84 & &.42 & .39 & .4  \\ 
        & gold  	& .91    & .85    & .88 & \multicolumn{3}{c}{.90} & & \multicolumn{3}{c}{.50} \\   \midrule
       \maskedmodel & predicted & .78	 & .76	 & .77 & .86	 & .82 & .84 & &  .69 & .69 & .69\\ 
         & gold  & .91    & .86    & .88 & \multicolumn{3}{c}{.91} & & \multicolumn{3}{c}{\textbf{.74}}\\   \bottomrule
    \end{tabular}
    \caption{Results on OntoNotes test data (English): document-level coreference resolution (only with unmasked mentions; CoNLL scores) and antecedent prediction (both unmasked and masked mentions); P, R, F1 = precision, recall, F1 scores (when using gold mention boundaries on antecedent prediction, P = R = F1). BUC, M$^3$ and CEAF scores are reported in Appendix \ref{appendix_b}.}
    \label{tab:coreference-results}
\end{table*}

Table~\ref{tab:coreference-results} reports the results of evaluation on OntoNotes test data for both \unmaskedmodel (the standard SpanBERT-coref coreference system) and our variant \maskedmodel, trained with 15\% of mentions masked in each document.\footnote{
    Models trained masking between 10\%-35\% of mentions achieved comparable antecedent accuracy on masked mentions on development data.
    Among the best setups, we select for analysis the best model in terms of coreference resolution.}
The table reports results for both model-predicted and gold mention boundaries; the latter are always higher, as expected.
For unmasked mentions, we provide results both for standard coreference resolution and per-mention antecedent predictions (\textsc{Antecedent}); for masked mentions, only the latter is applicable (see Section~\ref{sec:method}).

On unmasked mentions, the two models perform basically the same.
This means that masking 15\% of mentions during training, which was done only for \maskedmodel, does not interfere with the ability of the system on ordinary, unmasked coreference resolution.
On masked mentions, both models perform worse, which is expected because it is a more difficult task: Without lexical information about the mention, the models have less information to base their prediction on.
Still, both models provide correct predictions in a non-trivial portion of cases.
\unmaskedmodel, which did not observe any masked mentions during training, achieves .5 F1 for gold mentions.
A random baseline gets only .08, and selecting always the immediately previous mention or always ``no antecedent'' obtain .23 and .26, respectively.
Thus, it seems that \unmaskedmodel retains some ability to compute meaningful representations for masked tokens from pretraining, despite not seeing \texttt{[MASK]} during training for coreference resolution.
Nevertheless, in line with our expectations, training on masked coreference is beneficial: \maskedmodel improves substantially over the results of \unmaskedmodel, with .74 F1.
This means that even without lexical information from the mention itself, 74\% of referents are correctly identified, i.e., predictable on the basis of context alone.

\begin{figure}[htb]
     \centering
     \includegraphics[width = 0.85\linewidth]{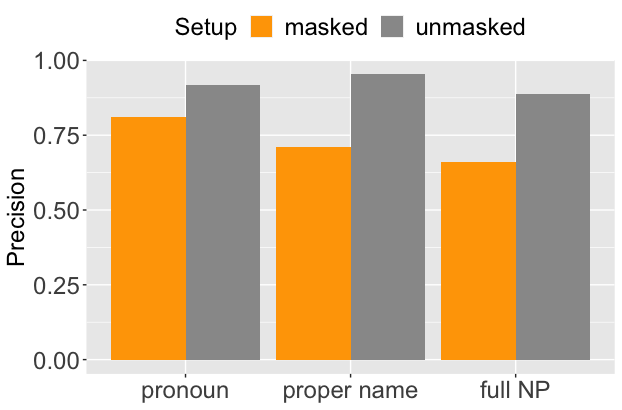}
     \caption{\label{fig:accuracy_fine_tuned} Antecedent precision for \maskedmodel across different mention types, for masked and unmasked mentions. }
\end{figure}

\paragraph{Results by mention type.}
Figure~\ref{fig:accuracy_fine_tuned} breaks down the antecedent precision scores of \maskedmodel by mention type.
From now on we look only at the setup with gold mention boundaries, though the trends are the same for predicted mentions (reported in Appendix \ref{appendix_b}).
We distinguish between proper names (e.g., ``Kamala Harris''), full noun phrases (NPs; e.g., ``the tall tree'') and pronouns (e.g., ``she'', ``my'', ``that'').
For completeness, in Appendix \ref{appendix_b} we report the results considering a more fine-grained distinction. 

The figure shows that for predicting the antecedent of masked mentions, pronouns are the easiest (.81), followed by proper names (.71), and full NPs (.66) are the hardest.
Put differently, pronouns are used in places where the referent is the most predictable, full NPs when the referent is the least predictable.
Table \ref{tab:examples} shows examples of predictions on masked mentions with different mention types. 

For unmasked mentions, instead, proper names are the easiest (.96; names are typically very informative of the intended referent), and full NPs (.89) are only slightly more difficult than pronouns (.92).
Hence, the pattern we see for masked mentions cannot be a mere side-effect of pronouns being easier to resolve in general (also when unmasked), which does not seem to be the case.
Instead, it provides initial evidence for the expected relation between referent predictability and mention choice, which we will investigate more in the next section.

\paragraph{Comparison to human predictions.}
We assess how human-like our model \maskedmodel behaves by comparing its outputs to human guesses in the cloze-task data from \citet{modi2017modeling}. 
Subjects were asked to guess the antecedent of a masked mention in narrative stories while seeing only the left context (182 stories, $\sim$3K mentions with 20 guesses each). 
To evaluate the model's estimates, we follow Modi et al.'s approach, and compute Jensen-Shannon divergence to measure the dissimilarity between a model's output and the human distribution over guessed referents (the lower the better).
\maskedmodel achieves a divergence of .46, better than Modi et al.'s best model (.50), indicating that our system better approximates human expectations. 
Appendix \ref{appendix_b} provides further results and details.

\section{Predictability and mention form}
\label{sec:referential_choice}

\begin{table*}[]
    \centering
    \footnotesize
    \begin{tabular}{p{14cm}p{.8cm}} \toprule
        context & mention \\ \toprule
        \footnotesize{
        (1) Judy Miller is protecting another source [...]
        Let me get a response from Lucy Dalglish. I think it's very obvious from what \dotuline{\underline{Judy}} wrote today \textbf{[MASK]} is protecting somebody else.} \checkmark & she   \\ \midrule
        \footnotesize{(2) This child [...]
        felt particularly lonely and especially wanted \underline{his father} to come back. He said that \dotuline{he} was sick one time. \textbf{[MASK]} worked in Guyuan}  $\times$ & his father   \\ \midrule
        \footnotesize{(3) One high-profile provision [...] 
         was the proposal by \dotuline{\underline{Chairman Lloyd Bentsen 
         of the Senate Finance Committee}} to expand the deduction for individual retirement accounts. \textbf{[MASK]} said he hopes the Senate will consider that measure soon}  \checkmark & Mr. Bentsen   \\ \midrule
        \footnotesize{(4) 
        \underline{Sharon Osbourne, Ozzy's long-time manager, wife and best friend}, announced to the world that she'd been diagnosed with colon cancer. Every fiber of \dotuline{Ozzy} was shaken. \textbf{[MASK]} had to be sedated for a while.} $\times$ & he   \\
        \bottomrule
    \end{tabular}
    \caption{Examples of correct and incorrect predictions by \maskedmodel (with gold mention boundaries) on masked mentions; model's prediction underlined, correct antecedent with dotted line. 
    }
    \label{tab:examples}
\end{table*}

The previous section assessed the effect of our masked training method on model quality. 
We believe that the model predictions are of high quality enough that we can use them to test the main hypothesis regarding the relation between predictability and mention choice.
Following previous work (see Section~\ref{sec:relatedwork}), we define predictability in terms of the information-theoretic notion of \textbf{surprisal}: the more predictable an event, the lower our surprisal when it actually occurs.
Given a masked mention $x$ with its true referent $e_{\text{true}}$, surprisal is computed from the model's output probability distribution over entities $E_x$ (Eq. \ref{prob_entity}), given the context $c_x$:

\begin{equation*}
    \text{surprisal}(x) := -\log_{2} P(E_x = e_{\text{true}} \mid c_x)
\end{equation*}
Surprisal ranges from 0 (if the probability assigned to the correct entity equals 1) to infinity (if this probability equals 0). 
Surprisal depends only on the probability assigned to the correct entity, regardless of the level of uncertainty between the competitors.
As \citet{tily2009refer} note, uncertainty between competitors is expected to be relevant for mention choice, e.g., a pronoun may be safely used if no competitors are likely, but risks being ambiguous if a second entity is somewhat likely.
\citet{tily2009refer} and, following them, \citet{modi2017modeling} took this uncertainty into account in terms of entropy, i.e., \emph{expected} surprisal.
We report our analyses using entropy in Appendix \ref{appendix_c}, for reasons of space and because they support essentially the same conclusions as the analyses using just surprisal.

We check whether surprisal predicts mention type (Section~\ref{sec:surprisalmentiontype}) and whether it predicts mention length (number of tokens; Section~\ref{sec:surprisalmentionlength}).
All analyses in this section use the probabilities computed by \maskedmodel with gold mention boundaries. 

\subsection{Surprisal as a predictor of mention type}
\label{sec:surprisalmentiontype}
\begin{figure}[!htb]
    \centering
     \includegraphics[width=0.8\linewidth]{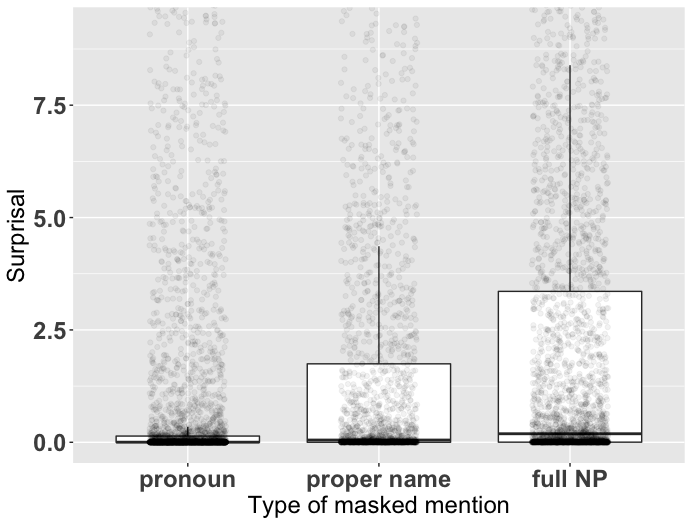}
     \caption{\label{fig:boxplot_surprisal} Surprisal and mention type. The limits on the y axis were scaled to the 95th percentile of the data to visualize the variability better.}
\end{figure}

For this analysis, in line with previous studies, we consider only third person pronouns, proper names and full NPs with an antecedent (i.e., not the first mention of an entity). 
For the OntoNotes test data this amounts to 9758 datapoints (4281 pronouns, 2213 proper names and 3264 full NPs).
Figure~\ref{fig:boxplot_surprisal} visualizes surprisal of masked mentions grouped by type,
showing that despite much within-type variation, full NPs tend to have higher surprisal (be less predictable) than pronouns and proper names.

To quantify the effect of predictability on mention type, we use multinomial logistic regression, using as the dependent variable the three-way referential choice with pronoun as the base level, and surprisal as independent variable.%
\footnote{We use the \textit{multinom} procedure from the library \textit{nnet} \citep{Venables2002nnet}. Continuous predictors were standardised, thus allowing for comparison of coefficients.}
The results of this surprisal-only regression are given in the top left segment of Table~\ref{tab: regression_summary}.
The coefficients show that greater surprisal is associated with a higher probability assigned to proper names ($\beta = .31$) and even more so full NPs ($\beta = .47$); hence pronouns are used for more predictable referents.
Since surprisal was standardized, we can interpret the coefficients (from logits to probabilities): e.g., adding one standard deviation from mean surprisal increases the predicted probability of a proper name from .23 to .25, and of a full NP from .33 to .42, decreasing the probability of a pronoun from .43 to .34.

Next, following \citealt{tily2009refer, modi2017modeling}, we test whether predictability has any effect over and above shallower linguistic features from the literature that have been hypothesized to affect mention choice.
We fit a new regression model including the following features as independent variables alongside surprisal:\footnote{
    The result of this simultaneous regression as regards the predictor surprisal will be identical to what the result would be of a hierarchical regression where surprisal is the last added predictor \citep{wurm2014residualizing}. 
} 
\textbf{distance} (num.\ sentences between target mention and its closest antecedent); \textbf{frequency} (num.\ mentions of the target mention's referent so far); closest \textbf{antecedent is previous subject} (i.e., of the previous clause);
\textbf{target mention is subject}; closest \textbf{antecedent type} (pronoun, proper name, or full NP).
The results are shown in the bottom left segment of Table~\ref{tab: regression_summary}.\footnote{
        Because the features themselves may capture aspects of predictability, and are indeed correlated with predictability (though all \textit{r} < .35), we cannot in this case interpret the coefficient for surprisal as directly indicative of the magnitude of the effect of predictability on mention choice.
    We visualize the comparison of observed to predicted types using a ternary plot, see Figure 6 in Appendix \ref{appendix_c}.}
We verified that the incorporation of each predictor improved goodness-of-fit, using the Likelihood Ratio (LR) chi-squared test (with standard .05 alpha level; see Appendix \ref{appendix_c} for specific results).
Surprisal improved goodness-of-fit ($p_{\chi^2} << 0.001$): it contributes relevant information not captured by the shallow features alone.
At the same time, however, now surprisal is not anymore predictive of the distinction between pronouns and proper names, as found by \citet{tily2009refer} -- only of the distinction between pronouns and full NPs (see significance values of the predictor `surprisal' for the two left columns of Table~\ref{tab: regression_summary}).

If we conceive of the shallow features as possible confounds, our results shows that predictability still affects mention choice when controlling for these. 
Alternatively, we can take the shallow features to themselves capture aspects of predictability (e.g., grammatical subjects tend to be used for topical referents, which are therefore expected to be mentioned again), in which case the results show that these features do not capture all aspects.

As for the shallow features themselves, we find that pronouns are favoured over proper names and full NPs when the referent has been mentioned recently, in line with the idea that the use of pronouns is sensitive to the local salience of a referent.
Moreover, pronominalization is more likely if the previous mention of the referent was itself a pronoun. 
There is also a strong tendency to reuse proper names, perhaps due to stylistic features of the texts in OntoNotes: in news texts, proper names are often repeatedly used, plausibly to avoid confusion, as news articles often introduce many entities in a short span; in the Bible, the use of repeated proper names is especially common for the protagonists (e.g.\ Jesus).
Lastly, we find the well-known \emph{subject bias} for pronouns: pronouns are more likely than full NPs or proper names when the referent's previous mention occurred in subject position.

Overall, the results corroborate the finding in \citet{tily2009refer} that full NPs are favoured, and pronouns and proper names disfavored, when surprisal is higher; 
and extend their finding, based on newspaper texts only, to a larger amount of data and more diverse genres of text (news, magazine articles, weblogs, religious texts, broadcast and telephone conversation).

\begin{table*}[h]
    \small
    \centering
    \begin{tabular}{ll|llll|llll|llll} 
  
         &                           & \multicolumn{8}{c|}{Predicting mention type}      &  \multicolumn{4}{c}{Predicting mention length}   \\ \hline   
         &                           & \multicolumn{4}{c|}{Proper name}             &\multicolumn{4}{c|}{Full NP}  &  & &  \\ 
         &                           & $\beta$  & s.e.  & $z$  & \textit{p}                       & $\beta$  & s.e.  & $z$  & \textit{p}            & $\beta$  &  s.e.  &  $t$ & \textit{p} \\ \hline 
\multicolumn{2}{l|}{Intercept}       &   -.63 & .03   & -23.8  & -                     &   -.26     &   .02    &  -10.9  & -  &  1.87  &  .02  &  80.8   & -   \\       
\multicolumn{2}{l|}{surprisal}      &  .31  & .03   &  9.6   & *                       &  .47    &   .03    &  16.4  & *   &  .25  &  .02  &  10.7  & * \\  \hline        
\multicolumn{2}{l|}{Intercept}       &   -.24 & .07   & -3.6   & -                     &   .04     &   .07    &  .6  & -    &  1.81  &  .05  &  40.1  & -   \\       
\multicolumn{2}{l|}{distance}      &  3.13  & .12   &  25.4   & *                       &    3.10   &   .12    &  25.2  & *   &   .17  &  .02  &  7.1  & *\\ 
\multicolumn{2}{l|}{frequency}       &  .09   &  .03    &  3.1  & *                     &   -.13     &   .03    &   -3.8  & *  &  -.13    &  .02  & -5.4  & *          \\
antecedent & previous subject       & -1.31 &  .09  &  -13.9  & *                      &   -1.10    &   .08      &  -13.7  & *     &  -.51    &  .06  &  -8.5   & *   \\
mention & subject               &  .07   &  .07  & 1.0    &  .3                          &   -0.50     &  .06     &  -7.7 & *    &  .04  &  .05  &  .8 &  .4 \\
antecedent type & proper name   &   1.78  & .08  & 22.8 & *                             &   .41      &   .09     &   4.6    & *     &  -.21   &  .06  & -3.2 & * \\
                 &full NP        &    -.17  &   .08       &  -2.2 & *                 &   1.18    & .06       &  18.1   & *        &  .42  &  .06 &  7.5 & * \\
\multicolumn{2}{l|}{surprisal}   &     .05   &   .04      &   1.5   & .1              &   .23     &  .03      &  7.8 & *         &  .17    &  .02  &   7.4  & * \\

\hline
    \end{tabular}
    \caption{(left) Two Multinomial logit models predicting mention type (baseline level is ``pronoun"), (right) two linear regression models predicting mention length (number of tokens) of the masked mention, based on 1) surprisal alone and 2) shallow linguistic features + surprisal. * marks predictors that are significant at the .05 alpha level. All predictors in models improved goodness-of-fit to the data except ``target mention is subject'' in the fuller linear regression model. Full tables with Likelihood-ratio Chi-squared test and F-test are reported in Appendix \ref{appendix_c}. }\label{tab: regression_summary}
\end{table*}

\subsection{Surprisal as a predictor of mention length}
\label{sec:surprisalmentionlength}

If pronouns are favoured for more predictable referents due to a trade-off between information content and cost, one would expect to find similar patterns using graded measures of utterance cost, instead of flattening it to coarse-grained distinctions across mention types.
In this subsection we use the number of tokens as such a measure \citep{orita2015discourse}.
The average number of tokens per mention in our dataset is (of course) 1 for pronouns, 1.67 for proper names and 3.16 for full NPs.

\begin{figure}[!htb]
    \centering
     \includegraphics[width=1\linewidth]{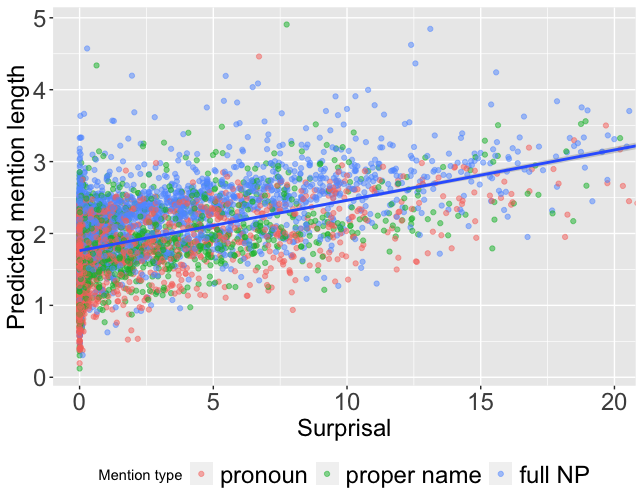}
     \caption{\label{fig: linear_length_surprisal} Trend between surprisal and predicted mention length by the linear regression model, visualized by adding a smoothing line comparing only the outcome with the variable surprisal.}
\end{figure}

We fit linear regression models with \textit{mention length} in number of tokens as the dependent variable (or number of characters, in Appendix \ref{appendix_c}), and, again, surprisal with and without shallow linguistic features as independent variables.
The right segment of Table~\ref{tab: regression_summary} presents the results, indeed showing an effect of mention length.
In the surprisal-only model, moving up by one standard deviation increases the predicted mention length by .25 tokens (or 1.40 characters, see Table 5 in Appendix \ref{appendix_c}).
Grammatical function and type of the antecedent are still strong predictors, with surprisal again making a contribution on top of that: mentions that refer to a more surprising referent tend to have more words.
Figure~\ref{fig: linear_length_surprisal} visualizes this trend between surprisal and predicted mention length.

Single-token pronouns dominate the lower end of the output range, raising the question of whether predictability still makes a difference if we exclude them, i.e., fit regression models only on the non-pronominal mentions.
Our results support an affirmative answer (see Table 4 and 6 in Appendix \ref{appendix_c}): the more surprising a referent, the longer the proper name or full NP tends to be.

\section{Discussion and Conclusion}

In this work, we studied the relationship between referent predictability and the choice of referring expression using computational estimates of the former. 
To derive these, we adapted an existing coreference resolution system to operate in a setup resembling those of cloze tasks employed in psycholinguistics. 
Using computational estimates of semantic expectations allowed us to scale and expedite analyses on a large dataset, spanning different genres and domains.

Our results point to a trade-off between clarity and cost, whereby shorter and possibly more ambiguous expressions are used in contexts where the referent is more predictable. 
We found this both when grouping mentions by their morphosyntactic type and by their length.
Referent predictability seems to play a partially overlapping but complementary role on referential choice with features affecting the salience of an entity, such as its recency, frequency or whether it was last mentioned as a subject.
This points to the open question as to whether salience can actually be reduced to predictability \cite{arnold2001effect, zarcone2016salience}.

Our bidirectional setup is not directly comparable to that of some of the related work as to the amount and scope of context given for prediction. Referents are predicted with only the preceding context in previous work, both in psycholinguistic and computational approaches, while our model gives predictions based on both the preceding and following contexts. 
If one important hypothesis both previous studies and our study aim at testing is that speakers tend to avoid the redundancies between the informativeness of context and that of referring expression, our results then point to an issue that merits more attention: 
What kind of context influences referential choice? Is it only the preceding one, or the following one? How much (on either side)?
\citet{leventhal1973effect} raised a similar question concerning word intelligibility in sentences and found that participants delayed the decision about a test word presented in noise until the entire sentence was encoded, and that the context after the target word was more facilitating to its intelligibility.
\citet{song2020forward} also showed that comprenhenders actively utilized post-pronominal information in pronoun resolution.
The use of a computational model provides flexibility to compare predictions using different amounts of context, and could shed light on how the previous and following context affect mention choice. 
Future work could also use unidirectional models, which allow for a setup more like the one adopted by prior work for ease of comparison, if requirements on the quality of performance can be met.

We hope that our work will foster the use of computational models in the study of referential choice.
Our methodology can be applied to more languages besides English (provided the availability of coreference resources; for instance, Arabic and Chinese are included in OntoNotes 5.0) and the study of phenomena beyond those considered here.
Relevant future venues are more fine-grained classifications of NPs (such as indefinite vs.\ definite), the effect of referent predictability on processing \citep{mcdonald1995time}, and the kinds of context examined in psycholinguistic experiments (e.g., different rhetorical relations, verbs with contrasting referent biases; \citealt{rohde2014grammatical,mayol2018asymmetries}).

\section*{Acknowledgements}
We thank Thomas Brochhagen, Andreas Mädebach and Laia Mayol for their valuable comments. 
This project has received funding from the European Research Council (ERC) under the European Union's Horizon 2020 research and innovation programme (grant agreement No. 715154). This paper reflects the authors' view only, and the EU is not responsible for any use that may be made of the information it contains. Matthijs Westera also received funding from Leiden University (LUCL, SAILS). We are grateful to the NVIDIA Corporation for the donation of GPUs used for this research. 

\bibliographystyle{acl_natbib}
\bibliography{main}
\section*{Appendices}

\appendix
\section{Method: details}
\label{appendix_a}

For simplicity, both in training and evaluation, we never mask mentions which are embedded in another mention (e.g., ``the bride'' in ``the mother of the bride''), since that would cover information relevant to the larger mention.
In case we mask a mention that includes another mention, we discard the latter from the set of mentions for which to compute a prediction.
    
For evaluation on development data, to find the best models across training epochs and hyperparameters, we use a quicker but more coarse-grained method than that used for evaluation on test data to assess performances on masked mentions.
We mask a random sample (10\%; independently of the percentage used during training) of mentions in each document, compute evaluation scores and get the average of these across 5 iterations (i.e., with different samples of mentions masked).
Although in this setup masks could potentially interfere with each other, and we will not have masked predictions for all mentions,
overall this method will give us a good enough representation of the model's performances on masked mentions, while being quick to compute.  

When evaluating antecedent prediction, we skip the first mention in a document as this is a trivial prediction (no antecedent).

\section{Evaluation}
\label{appendix_b}
\subsection{Complete results on OntoNotes}

Table \ref{tab:coreference-scores} reports MUC, B$^3$ and CEAF scores (precision, recall and F1), for the \unmaskedmodel and \maskedmodel. 
The results are overall comparable between the two systems across all metrics.

\begin{figure}[!htb]
    \centering
     \includegraphics[width=1\linewidth]{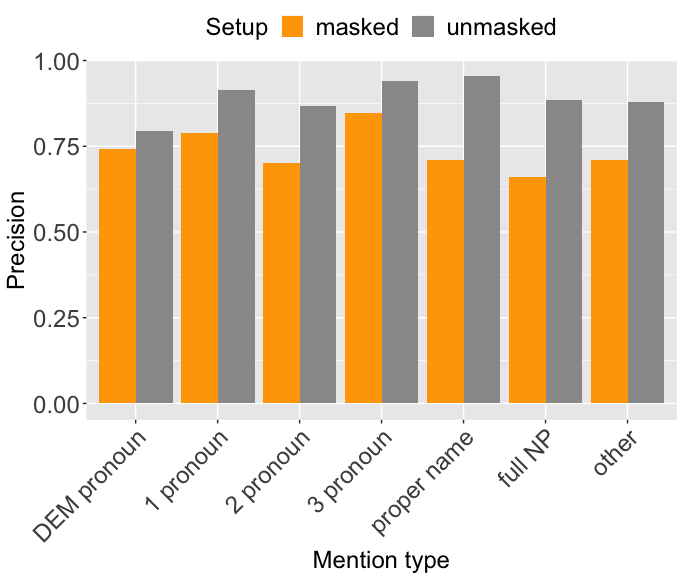}
     \caption{Antecedent precision for \maskedmodel across more fine-grained mention types, for masked and unmasked mentions.}\label{fig:masked_fine_tuned_finer_grained}
\end{figure}

\begin{figure}[!htb]
    \centering
     \includegraphics[width=1\linewidth]{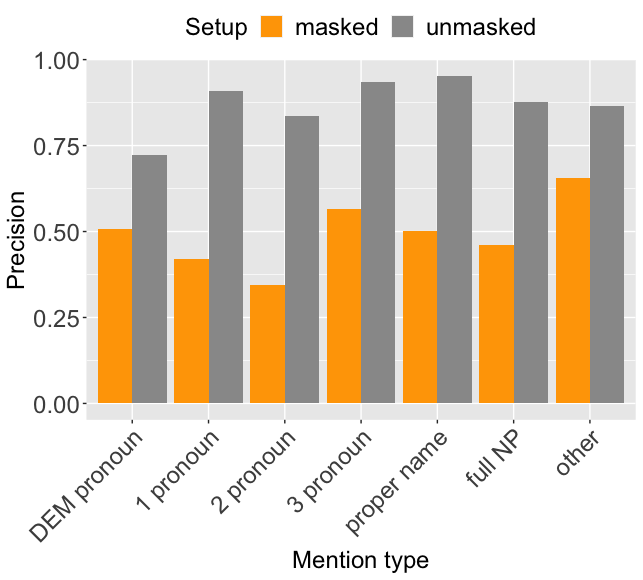}
     \caption{Antecedent precision scores with gold mentions of the model \unmaskedmodel across different mention types, for both masked and unmasked mentions.}\label{fig:masked_original_finer_grained}
\end{figure}

Figures \ref{fig:masked_fine_tuned_finer_grained} and \ref{fig:masked_original_finer_grained} report the antecedent prediction results, using gold mention boundaries, of the \maskedmodel and \unmaskedmodel considering a fine-grained distinctions across mention types than what reported in Figure 2 of the paper.
Concretely, we divide pronouns, into first-, second- and third-person pronouns, as well as treating demonstratives (e.g., ``that'') as a separate category (DEM).
We subdivide pronouns in this way because they are quite heterogeneous: first- and second-person pronouns are comparatively rigid (typically referring to the speaker and addressee), and are used oftentimes within a quotation (e.g. 
Asked why \dotuline{\underline{senators}} were giving up so much, New Mexico Sen. Pete Dominici, [...] said, ``[\textbf{We}]'re looking like idiots [...]");
and demonstrative pronouns tend to be more difficult cases in OntoNotes, for instance referring to the head of verbal phrases (e.g. 
[...] their material life will no doubt be a lot less \dotuline{\underline{taken}} of when compared to the usual both parents or one parent at home situation. [\textbf{This}] is indisputable).
Overall, for masked mentions, precision is high across pronouns, and highest among pronoun types for third-person pronouns. 
For unmasked mentions, the hardest cases are demonstrative pronouns. 

We also report these results looking at predictions with predicted (i.e., identified by the system) mention boundaries. These are displayed in Figure \ref{fig:accuracy_original_predicted_mentions} and \ref{fig:accuracy_fine_tuned_predicted_mentions} for \unmaskedmodel and \maskedmodel, respectively. 
While results are generally better with gold mention boundaries, the trends stay the same across the two setups for both masked and unmasked mentions. 

\begin{figure}[!htb]
    \centering
     \includegraphics[width=\linewidth]{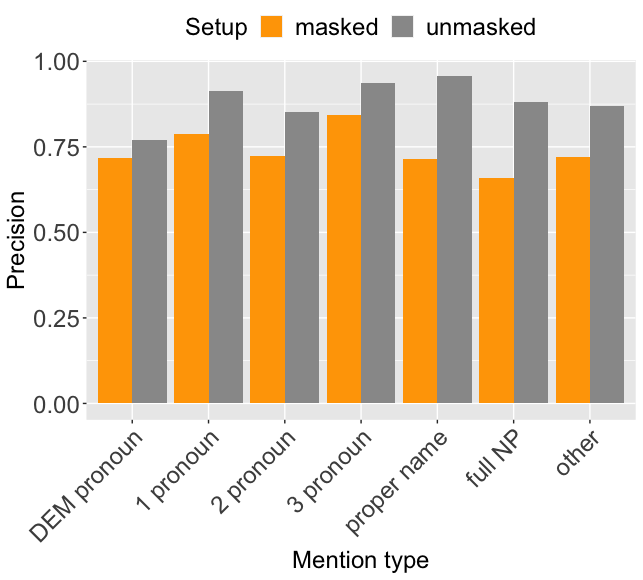}
     \caption{Antecedent precision comparison of masked cloze task across mention types, for both masked and unmasked mentions, with three [MASK] tokens.}\label{fig:masked_accuracy_three_masks}
\end{figure}

\begin{figure*}[!htb]
   \begin{minipage}{0.48\textwidth}
     \includegraphics[width=1\linewidth]{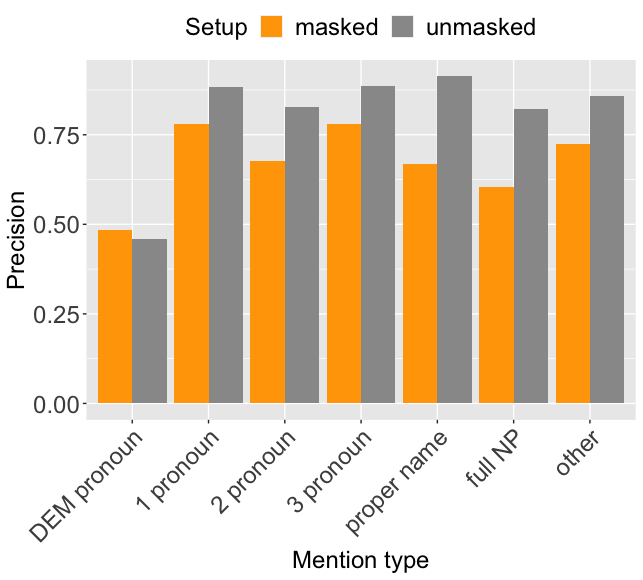}
     \caption{Antecedent precision scores with predicted mentions of the model \maskedmodel across different mention types, for both masked and unmasked mentions.}\label{fig:accuracy_fine_tuned_predicted_mentions}
   \end{minipage}\hfill
   \begin{minipage}{0.48\textwidth}
     \includegraphics[width=1\linewidth]{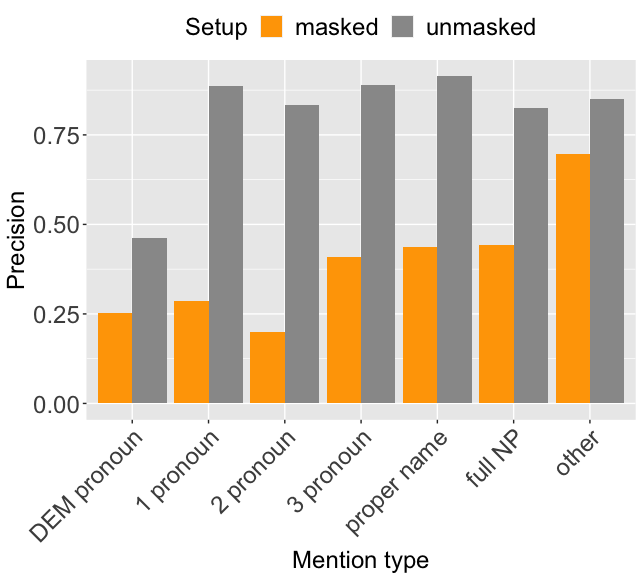}
     \caption{Antecedent precision scores with predicted mentions of the model \unmaskedmodel across different mention types, for both masked and unmasked mentions.}\label{fig:accuracy_original_predicted_mentions}
   \end{minipage}
\end{figure*}

Finally, in Figure \ref{fig:masked_accuracy_three_masks} we report the results looking at a variant of \maskedmodel where instead of substituting mentions with one \texttt{[MASK]} token we use a sequence of three. 
This is to verify whether the use of a single token biases the system to be better on one-token mentions. 
The results show that this is not the case, as the trends found with the one-token masking are the same as those with the three-tokens masking: In particular, when a third-person pronoun is used the antecedent is still easier to predict than when a proper name is used, and even less than a full NP.

\begin{table*}[h]
    \centering
    \begin{tabular}{llllllllllllll} \toprule
        model & mentions & \multicolumn{3}{c}{MUC} & \multicolumn{3}{c}{B$^3$} & \multicolumn{3}{c}{CEAF} & \multicolumn{3}{c}{Average of metrics} \\
        & & P & R & F1 & P & R & F1 & P & R & F1 & P & R & F1 \\ \toprule
        \unmaskedmodel & predicted & .84 & .83	&  .84 & .76	 & .75	 & .76 & .75 &  .71 & .73  & .78	 & .77	 & .77 \\ 
                & gold  & .95 & .91	&  .93 & .87	 & .86	 & .86 &  .92 & .77  & .84	& .91    & .85    & .88 \\   \midrule
        \maskedmodel & predicted & .84 & .83	&  .83 & .76	 & .75	 & .75 & .74 &  .71 & .73  & .78	 & .76	 & .77 \\ 
                & gold  & .95 & .93	&  .94 & .86	 & .88	 & .87 &  .92 & .78  & .85	& .91    & .86    & .88 \\   \bottomrule
    \end{tabular}
    \caption{Results on OntoNotes test data (English) in document-level coreference resolution (only with unmasked mentions); P, R, F1 = precision, recall, F1 scores. }
    \label{tab:coreference-scores}
\end{table*}

\subsection{Comparison to human predictions}

To elicit human judgments of referent predictability, \citet{modi2017modeling} relied on mention heads rather than the complete mention (e.g., ``supermarket'' in ``the supermarket'').
For one, they constructed the cloze task by cutting a text right before the head of the target mention (e.g., before ``supermarket''), thus leaving part of the mention visible (e.g., ``the'' in this case).
Moreover, they indicated candidate antecedent mentions for the human participants to consider, by listing again only the mention heads.

To make this task suitable for standard coreference resolution we need to identify the full mention boundaries belonging to each head (not given in the original annotations).
To that end we rely on `noun chunks' identified by the \href{https://spacy.io/usage/linguistic-features#noun-chunks}{spaCy} library, amended by a number of heuristics, for an estimated 91\% accuracy (estimated by manually checking a sample of 200 mentions for correctness).
We use the identified mention boundaries as gold mention boundaries exactly as in our OntoNotes setup (Section 3).
However, different from our OntoNotes setup, we mask only the head of the target mention, exactly as in the human cloze task.

Table~\ref{tab:human_model_alignment} reports the results of \maskedmodel on the data by \citet{modi2017modeling}. 
We deploy the system in two setups: 1) Using just the left context of the target mention, mimicking the setup used to elicit the human judgments, and 2) Using both the left and right context of the mention.
In both cases, our results improve over those reported by \citet{modi2017modeling} for their best model, indicating that through our method we obtain better proxies for human discourse expectations. 

\maskedmodel's predictions are more aligned to those of humans when accessing both sides of the context than with only the left context, in spite of the second setup more closely resembling that used for the human data collection.
Since information in the following context could not influence the human judgements (it was not available), we take this result to indicate that \maskedmodel works generally better when deployed in a setup that is closer to that used during its training (recall that in training it never observed texts cropped after a masked mention), leading to suboptimal predictions when only the left context is used. 
We plan to explore this further in future work, by experimenting with variants to the training setup or different architectures (e.g., auto-regressive) that may improve the model's ability to resolve mentions based only on their previous contexts.

\begin{table*}[h]
    \centering
    \begin{tabular}{lccc} 
  
        &    accuracy  & relative accuracy w.r.t human top guess    & JSD         \\ \toprule
   \maskedmodel left only    &    .54            & .50                                             & .46      \\
  \maskedmodel left + right  &    .74            & .64                                             & .39      \\ \midrule
  \citet{modi2017modeling}  & .62 &    .53            & .50     \\
\hline
    \end{tabular}
    \caption{Evaluation of \maskedmodel against human guesses using different amounts of context, in terms of average relative accuracy with respect to human top guess, as well as average Jensen-Shannon divergence (smaller is better) between the probability distribution of human predictions and model predictions.}
    \label{tab:human_model_alignment}
\end{table*}

\section{Predictability and mention form}
\label{appendix_c}

\subsection{Regression with both surprisal and entropy}

In addition to surprisal, \citet{tily2009refer} and \citet{modi2017modeling} also consider the uncertainty over competitors as a feature that captures some aspect of predictability. 
This uncertainty, more precisely \textbf{entropy}, is defined as \emph{expected} surprisal:
\begin{equation*}
\text{entropy}(x) := \sum_{e \in E_x} P(E_x = e \mid c_x) \cdot \text{surprisal}(x) 
\end{equation*}
Entropy will be low if the probability mass centers on one or a few entities, and high if the probability is more evenly distributed over many entities, regardless of which entity is the correct one.

In principle, entropy and surprisal capture genuinely different aspects of predictability; for instance, when the model is confidently incorrect, surprisal is high while entropy is low.
However, in our data, entropy and surprisal are highly correlated ($\textit{r}_{s}$ = .87, \textit{p} $<$ .001). 
We did not fit regression models with both by residualising entropy to eliminate the collinearity, as our precedents did, because of the shortcomings of treating residualisation as remedy for collinearity \citep{wurm2014residualizing}. 
Instead, we define predictability primarily by surprisal (Uniform Information Density, \citealt{levy2007speakers}) in our main analysis, and report the regression with both surprisal and the non-residualised entropy as a supplementary analysis. 
Note that we do not intend to interpret the coefficient of surprisal or entropy in this analysis (this is not possible because they are collinear), but rather to test whether surprisal and entropy still improve goodness-of-fit to the data on top of many other shallow linguistic features.
Again, the shallow features themselves may capture aspects of entropy, and are indeed correlated with entropy (though all $r < .50$). 

Table \ref{tab: multinomial_type_surprisal_entropy} shows that both surprisal and entropy still matter for mention choice when controlling for the other factors, even though their statistical significance might be undermined due to the collinearity between them.
Compared to the model with predictability primarily formulated as surprisal, similar effect patterns are found with entropy added, except that entropy seems to be better at distinguishing between pronoun vs. non-pronouns, and as the contexts become more uncertain, proper names and full NPs are roughly equally favored ($z = -.88$, $p = .38$) over pronouns after controlling for other variables.

\begin{table*}[h]
    \centering
    \small
    \begin{tabular}{ll|llll|llll|lll} 
  
         &                           & \multicolumn{4}{c}{Proper name}             &\multicolumn{4}{c|}{Full NP}   &  & & \\ 
         &                           & $\beta$  & s.e.  & $z$   & \emph{p}                      & $\beta$  & s.e.  & $z$  & \emph{p}                 & $\text{LR}_{\chi^2}$   & df  &  $\text{\emph{p}}_{\chi^2}$         \\ \hline \hline
\multicolumn{2}{l|}{Intercept}       &   -.61 & .03   & -22.8   & -                    &   -.25     &   .02    &  -10.3  & -              &    &   &     \\       
\multicolumn{2}{l|}{surprisal}      &  .16  & .03   &  5.1   & *                      & .33    &   .03    &  12.4  & *   &   330.3  &  2 &  *                     \\  
\multicolumn{2}{l|}{entropy}   &     .42   &   .03  &   14.3   & *                   &  .45    &    .03.  & 16.8   & * &  349.4  & 2  &  *                    \\ \hline \hline         
\multicolumn{2}{l|}{Intercept}       &   -.29 & .07   & -4.2   & -                     &   -.01     &   .07    &  -0.2  & -              &    &   &     \\       
\multicolumn{2}{l|}{distance}      &  3.01  & .12   &  24.4  &  *                      &    3.00    &   .12    &  23.1  & *   &   1294.3  &  2 & *                     \\ 
\multicolumn{2}{l|}{frequency}       &  .09   &  .03    &  3.3  & *                     &   -.12     &   .03    &   -3.6  & *  &  37.1  &  2 &     *              \\
antecedent & previous subject   & -1.29 &  .09  &  -13.7  & *               &   -1.06       &   .08      &  -13.4  & *  &   346.6  &  2 &    *          \\
mention & subject    &  .12     &  .07  & 1.7 & .1               &   -0.44     &  .07     &  -6.8  & *    & 72.9   &  2 &   *                 \\
antecedent type & proper name    &   1.79   &      .08    &   22.8   & *                 &   .42    &   .09    &  4.7 & *  & \multirow{2}{2.5em}{1730.7}   & \multirow{2}{1em}{2}  & \multirow{2}{2.5em}{*}                  \\
                 &full NP        &    -.16  &   .08         &  -2.0  & *                &    1.20       & .07   &   18.3  & *  & &   &                   \\
\multicolumn{2}{l|}{surprisal}   &     -.01   &   .04         &   -0.2  & .8               &     .17  &    .03 &   6.0 & *  &  52.5  & 2  &  *                    \\
\multicolumn{2}{l|}{entropy}   &     .23   &   .03         &   6.7   & *              &     .25      &    .03      &  8.4 & *   &  77.5  & 2  &  *                    \\

\hline
    \end{tabular}
    \caption{Two Multinomial logit models predicting mention type (baseline level is ``pronoun"), based on 1) surprisal \& entropy and 2) shallow linguistic features + surprisal \& entropy. * marks predictors that are significant at the .05 alpha level.}\label{tab: multinomial_type_surprisal_entropy}
\end{table*}

\subsection{More analyses results}

\begin{figure}[!htb]
    \centering
     \includegraphics[scale=.37]{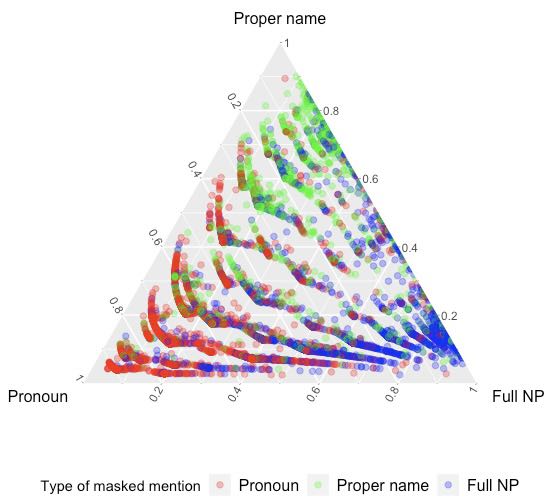}
   \caption{Ternary probability plot. Each point represents predicted probabilities from the multinomial logit model with shallow features and surprisal predicting the three levels of mention type, which sum to 1.}\label{fig:ternary_surprisal}
\end{figure}

Figure \ref{fig:ternary_surprisal} displays the predictions of mention type from the multinomial regression model, based on shallow features as well as surprisal. 
Each point represents a division of probability between the three levels of mention type. 
The corners of the triangle correspond to probability 1 for one outcome level and 0 for the other two, and the centre corresponds to probability 1/3 for each. 
Our model clusters most of the true pronouns (red) in the bottom left, and true full NPs (blue) in the bottom right, true proper names (green) at the top. 
Besides, many datapoints obtain similar division of probability, suggesting that some of them share similar pattern of features (recency, frequency etc.).

Table \ref{tab: linear_non_pronoun} shows two linear regression models predicting mention length quantified in terms of number of tokens for each non-pronominal mention (proper name, full NP). 
In the first model, we regress mention length (num. of tokens) on surprisal alone. 
In the second one, the coefficient of surprisal decreases a bit with other shallow linguistic features added.
F-tests are carried out to test if each predictor improves the fits to the data. 
In the fuller model, "frequency" and "antecedent type" are tested to significantly improve the model fit, above and over which surprisal still matters for mention type:
longer non-pronominal expressions are favoured with surprisal increasing.
We show similar effect pattern with two linear regression models predicting mention length alternatively measured in terms of characters, in Table \ref{tab:linear_character_count_non_pronoun}. 

Table \ref{tab: linear_character_count} displays two linear regression models predicting mention length measured in terms of number of characters for each masked mention (including pronominal mentions). 
Compared to models for non-pronominal mentions, features like "distance", "antecedent type" matter more when predicting the mention length with pronouns included, 
suggesting that these features better identify the distinction between pronouns vs. non-pronouns, but probably not between shorter and longer non-pronominal expressions.

Table \ref{tab: multinomial_type_surprisal_test} and \ref{tab: linear_length_surprisal_test} add results from likelihood-ratio chi-square tests and F-tests to Table 3 in the main text. All variables are tested to significantly improve goodness-of-fit to the data, except the feature "target mention is subject" in predicting mention length (num. of tokens).

\begin{table*}[h]
    \centering
    \begin{tabular}{ll|llll|lll}
                                  &         & $\beta$  &  s.e.  &  $t$   & $\text{\emph{p}}_{t}$  & F & df       &  $\text{\emph{p}}_{F}$ \\ \hline \hline
\multicolumn{2}{l|}{Intercept}              &  2.53  &  .04  &  64.24  &    -      &           &         &                         \\ \hline
\multicolumn{2}{l|}{surprisal}             &    .18 &   .04  &  5.10  &    *         & 26.00   &    1  &       *                       \\ \hline \hline                                  
\multicolumn{2}{l|}{Intercept}              &  2.69  &  .09  &  30.72  &    -      &           &         &                         \\ \hline
\multicolumn{2}{l|}{distance}             &  -.02  &   .03  &  -0.71   &  .48     &   .51   &    1      &     .48                       \\ \hline
\multicolumn{2}{l|}{frequency}              &  -.31    &  .05  &   -6.87  &   *    &  47.17    &    1      &   *                        \\ \hline
antecedent & previous subject &  -.14    &  .15  &  -.93   &   .35    &  .87   &     1     &    .35                       \\ \hline
mention & subject    &   .10  &  .09  &  1.05   &    .29      &  1.11  &      1    &       .29                      \\ \hline
antecedent type & proper name                &  -.96   &  .11  & -8.71   &    *     & \multirow{2}{2.5em}{77.26} &   \multirow{2}{1em}{2}       &  \multirow{2}{1em}{*}                         \\ 
              &  full NP                    &  .16  &  .10 &  1.56  &   .12    &            &          &                           \\ \hline
\multicolumn{2}{l|}{surprisal}              &  .16    &  .04  &   4.44  &   *    &   19.74   &    1      & *                  \\ \hline              
    \end{tabular}
    \caption{Two Linear regression models predicting mention length for each masked non-pronominal mention, based on 1) surprisal alone and 2) shallow linguistic features + surprisal. }\label{tab: linear_non_pronoun}
\end{table*}

\begin{table*}[h]
    \centering
    \begin{tabular}{ll|llll|lll}
                                  &         & $\beta$  &  s.e.  &  $t$   & $\text{\emph{p}}_{t}$  & F & df       &  $\text{\emph{p}}_{F}$ \\ \hline \hline
\multicolumn{2}{l|}{Intercept}              &  8.30  &  .11  &  75.68  &    -      &           &         &                         \\ \hline
\multicolumn{2}{l|}{surprisal}             &  1.40   &   .11  &  12.76  &    *         &  162.83  &    1  &       *                       \\ \hline \hline                                  
\multicolumn{2}{l|}{Intercept}              &  8.01  &  .21  &  37.68  &    -      &           &         &                         \\ \hline
\multicolumn{2}{l|}{distance}             &  1.00  &   .11  &  8.87   &  *     &   78.64   &    1      &     *                       \\ \hline
\multicolumn{2}{l|}{frequency}              &  -.79    &  .11  &   -6.93  &   *    &  48.04    &    1      &   *                        \\ \hline
antecedent & previous subject &  -3.02    &  .29  &  -10.58   &  *    &  111.88   &   1     &    *                       \\ \hline
mention & subject    &  -.02  &  .25  &  -.06   &    .95      &  .004  &      1    &       .95                      \\ \hline
antecedent type & proper name                &  -.40   &  .30  & -1.32   &  .19     & \multirow{2}{2.5em}{48.45} &   \multirow{2}{1em}{2}       &  \multirow{2}{1em}{*}                         \\ 
              &  full NP                    &  2.05  &  .26 &  7.86  &   *    &            &          &                           \\ \hline
\multicolumn{2}{l|}{surprisal}              &  .95    &  .11  &   8.57  &   *    &   73.52   &    1      & *                  \\ \hline              
    \end{tabular}
    \caption{Two Linear regression models predicting the character count (without space) for each masked mention, based on 1) surprisal alone and 2) shallow linguistic features + surprisal. F-test compares the fits of models.}\label{tab: linear_character_count}
\end{table*}

\begin{table*}[h]
    \centering
    \begin{tabular}{ll|llll|lll}
                                  &         & $\beta$  &  s.e.  &  $t$   & $\text{\emph{p}}_{t}$  & F & df       &  $\text{\emph{p}}_{F}$ \\ \hline \hline
\multicolumn{2}{l|}{Intercept}              &  12.19  &  .18  &  67.95  &    -      &           &         &                         \\ \hline
\multicolumn{2}{l|}{surprisal}             &  .88   &   .16  &  5.54  &    *        &  30.69 &    1  &       *                       \\ \hline \hline                                  
\multicolumn{2}{l|}{Intercept}              &  12.93  &  .40  &  32.47  &    -      &           &         &                         \\ \hline
\multicolumn{2}{l|}{distance}             &  -.08  &   .15  &  -.56   &  .58     &   .31   &    1      &     .58                       \\ \hline
\multicolumn{2}{l|}{frequency}              &  -1.68    &  .20  &   -8.18  &   *    &  66.83    &    1      &   *                        \\ \hline
antecedent & previous subject &  -.74    &  .67  &  -1.11   &  .27    &  1.24   &   1     &    .27                      \\ \hline
mention & subject    &  .63  &  .42  &  1.49   &    .14      &   2.21 &      1    &  .14                      \\ \hline
antecedent type & proper name                &  -4.16   &  .50  & -8.26   &  *     & \multirow{2}{2.5em}{63.40} &   \multirow{2}{1em}{2}       &  \multirow{2}{1em}{*}                         \\ 
              &  full NP                    &  .41  &  .47 &  .89  &   .37    &            &          &                           \\ \hline
\multicolumn{2}{l|}{surprisal}              &  .76    &  .16  &   4.75  &   *    &    22.58  &    1      & *                  \\ \hline              
    \end{tabular}
    \caption{Two Linear regression models predicting the character count (without space) for each masked non-pronominal mention, based on 1) surprisal alone and 2) shallow linguistic features + surprisal. F-test compares the fits of models.}\label{tab:linear_character_count_non_pronoun} 
\end{table*}

\begin{table*}[h]
    \centering
    \begin{tabular}{ll|lll|lll|llp{0.3cm}} 
  
         &                           & \multicolumn{3}{c}{Proper name}             &\multicolumn{3}{c|}{Full NP}   \\ 
         &                           & $\beta$  & s.e.  & $z$                         & $\beta$  & s.e.  & $z$                   & LR${{\chi}^2}$   & df  &  $p_{{\chi}^2}$ \\ \hline \hline
\multicolumn{2}{l|}{Intercept}       &   -.63 & .03   & -23.77                        &   -.26     &   .02    &  -10.93                &    &   &     \\       
\multicolumn{2}{l|}{surprisal}      &  .31  & .03   &  9.56                          &    .47    &   .03    &  16.44     &   330.28  &  2 &  *                   \\  \hline          
\multicolumn{2}{l|}{Intercept}       &   -.24 & .07   & -3.60                        &   .04     &   .07    &  0.58                &    &   &     \\       
\multicolumn{2}{l|}{distance}      &  3.13  & .12   &  25.40                          &    3.10    &   .12    &  25.23     &   1466.81  &  2 &  *                     \\ 
\multicolumn{2}{l|}{frequency}       &  .09   &  .03    &  3.11                       &   -.13       &   .03    &   -3.78    &  37.35  &  2 &    *             \\
antecedent & previous subject   & -1.31 &  .09  &  -13.91                &   -1.10       &   .08      &  -13.70    &   362.02  &  2 &     *        \\
mention & subject    &  .07     &  .07  & 1.00                &   -0.50     &  .06     &  -7.71      & 82.57   &  2 &   * \\
antecedent type & proper name    &   1.78   &      .08    &   22.83                    &   .41        &   .09     &   4.62    & \multirow{2}{2.5em}{1723.67}   & \multirow{2}{1em}{2}  & \multirow{2}{2.5em}{*}                  \\
                 &full NP        &    -.17  &   .08         &  -2.16                  &    1.18         & .06       &   18.13    & &                    \\
\multicolumn{2}{l|}{surprisal}   &     .05   &   .04         &   1.46                 &     .23      &    .03      &    7.80    &  75.18  & 2  &  *   \\

\hline
    \end{tabular}
    \caption{Two Multinomial logit models predicting mention type (baseline level is ``pronoun"), based on 1) surprisal alone and 2) shallow linguistic features + surprisal. Chi-square values of likelihood-ratio tests are indicative of any significant improvement in model by adding a predictor. $^*: p_{{\chi}^2} < 0.001$.}\label{tab: multinomial_type_surprisal_test}
\end{table*}

\begin{table*}[h]
    \centering
    \begin{tabular}{ll|llll|lll}
                                  &         & $\beta$  &  s.e.  &  $t$   & $\text{\emph{p}}_{t}$  & F & df       &  $\text{\emph{p}}_{F}$ \\ \hline \hline
\multicolumn{2}{l|}{Intercept}              &  1.87  &  .02  &  80.75  &    $^*$       &           &         &                         \\ 
\multicolumn{2}{l|}{surprisal}             &   0.25  &  0.02  &  10.74  &  $^*$      &   115.32   &    1      &     $^*$                        \\ \hline                          
\multicolumn{2}{l|}{Intercept}              &  1.81  &  .05  &  40.08  &    $^*$       &           &         &                         \\ 
\multicolumn{2}{l|}{distance}             &   0.17  &  0.02  &  7.08  &  $^*$      &   50.08   &    1      &     $^*$                        \\ 
\multicolumn{2}{l|}{frequency}              &  -.13    &  .02  &   -5.37  &   $^*$     &  28.82    &    1      &   $^*$                         \\ 
antecedent & previous subject               &  -.51    &  .06  &  -8.46   &   $^*$     &  71.59   &     1     &    $^*$                        \\ 
mention & subject                           &   .04  &  .05  &  .77   &    .44      &  .60  &      1    &       .44                      \\ 
antecedent type & proper name                &  -.21   &  .06  & -3.21   &  .0013      & \multirow{2}{2.5em}{59.32} &   \multirow{2}{1em}{2}       &  \multirow{2}{1em}{$^*$ }                         \\ 
              &  full NP                    &  .42  &  .06 &  7.51  &   $^*$     &            &          &                           \\ 
\multicolumn{2}{l|}{surprisal}              &  .17    &  .02  &   7.37  &   $^*$     &   54.37   &    1      &        $^*$                    \\ \hline              
    \end{tabular}
    \caption{\label{tab: linear_length_surprisal_test} Two Linear regression models predicting mention length (number of tokens) of the masked mention, based on 1) surprisal alone and 2) shallow linguistic features + surprisal.
    F-test compares the fits of nested models.
    All predictors were tested to improve goodness-of-fit to the data except "target mention is subject".  $^*: p < 0.001$}
\end{table*}

\end{document}